\documentclass[10pt, a4paper]{article}

\usepackage[final]{lrec2026} 
\usepackage{amsmath}
\usepackage{expex}
\usepackage{multirow}
\usepackage{tabularx}
\usepackage{array}
\usepackage{rotating}
\usepackage{natbib}
\usepackage{colortbl}
\usepackage[table]{xcolor}

\title{Is Semi-Automatic Transcription Useful in Corpus Creation? Preliminary Considerations on the KIParla Corpus}

\name{
\begin{tabular}{c}
Martina Simonotti$^{1}$, Ludovica Pannitto$^{1}$, Eleonora Zucchini$^{2}$\\
Silvia Ballarè$^{1}$, Caterina Mauri$^{1}$
\end{tabular}%
} 

\address{$^{1}$ University of Bologna, Bologna - Italy \\
         $^{2}$ Masaryk University, Brno - Czech Republic\\
         martina.simonotti@studio.unibo.it, ludovica.pannitto@unibo.it, \\ eleonora.zucchini@mail.muni.cz, silvia.ballare@unibo.it, caterina.mauri@unibo.it\\}

\abstract{
This paper analyses the implementation of Automatic Speech Recognition (ASR) into the transcription workflow of the KIParla corpus, a resource of spoken Italian. Through a two-phase experiment, 11 expert and novice transcribers produced both manual and ASR-assisted transcriptions of identical audio segments across three different types of conversation, which were subsequently analyzed through a combination of statistical modeling, word-level alignment and a series of annotation-based metrics. Results show that ASR-assisted workflows can increase transcription speed but do not consistently improve overall accuracy, with effects depending on multiple factors such as workflow configuration, conversation type and annotator experience. Analyses combining alignment-based metrics, descriptive statistics and statistical modeling provide a systematic framework to monitor transcription behavior across annotators and workflows. Despite limitations, ASR-assisted transcription, potentially supported by task-specific fine-tuning, could be  integrated into the KIParla transcription workflow to accelerate corpus creation without compromising transcription quality.
 \\ \newline  }

\begin{document}

\maketitleabstract

\section{Introduction}

As automatic speech recognition (henceforth, ASR) systems have become increasingly accurate, researchers have begun to question whether they could be effectively integrated into the construction pipeline of spoken resources. Manually transcribing spoken data is an extremely time-consuming task, with transcription-time to real-time ratios reaching up to 47 times the duration of the audio for the most challenging multilogues~\citep{Gorisch2024}. ASR-based transcription is also estimated to be twenty to forty times cheaper, given that manual transcription typically involves hiring professional transcribers~\citep{Umair2022}. In addition, proper training for transcribers is often required, and the transcription protocol must be both thorough, to ensure consistency, and practical, to be feasible for fully manual pipelines. 
Moreover, while transcription is often perceived as the most objective part of the resource creation process, there is ample evidence suggesting that even for languages with a solid written tradition, native speakers tend to paraphrase and regularize the signal~\citep{himmelmann2018transcription,jung2011retelling} with human error rates ranging from 3\% to 10\%, depending on the nature of the input speech and the amount of time dedicated to the transcription task~\citep{Ghyselen2020}.

In this paper, we describe the first steps taken towards the introduction of ASR models in the current workflow that supports the creation of the KIParla Corpus \citep{maurikiparla,ballare2020creazione}, a modular and growing resource of spoken Italian. 
The resource and current transcription workflow are described in Section~\ref{sec:kiparla}. Section~\ref{sec:literature} is devoted to a brief literature review on the topic. Section~\ref{sec:experiment} presents the experimental setup designed to evaluate the effect of introducing an ASR step into the workflow, while section~\ref{sec:results} provides preliminary results and considerations. Finally, Section~\ref{sec:conclusion} details limitations and future work\footnote{Data and code for this study are provided at \url{https://github.com/KIParla/asr-assisted-transcription}.}.

\section{The KIParla Corpus Methodology}
\label{sec:kiparla}

KIParla~\citep{maurikiparla,ballare2020creazione}\footnote{\url{https://kiparla.it/}} is a growing corpus of spoken Italian, the product of a collaborative effort between primarily the Universities of Bologna and Turin. 
At the moment of writing, it consists of around 228 hours of recording and approximately 2M transcribed tokens, and is freely available for consultation through aligned transcriptions, as well as full transcripts both in orthographical (i.e., following Italian standard orthography) and conversational format.
KIParla conversations are recorded in ecological settings, which range from university classes to participants' homes and can involve up to 5 speakers of various ages, genders, origins, occupations and education levels. Therefore, the corpus encompasses a diverse range of Italian spoken varieties and often capture non-standard linguistic features.  
Conversation types in the resource are also quite variable, including \textit{near-monologues} (recorded from university lectures), \textit{semi-structured interactions} (e.g., oral exams, office hours and interviews) as well as completely \textit{free conversations}. This variability significantly affects the number of conversational overlaps between speakers, which inevitably reduces the accuracy of off-the-shelf ASR models, as they typically struggle with speaker diarization. All conversations in KIParla are transcribed manually, to guarantee a high level of accuracy and allow special cases to be treated individually. The software adopted for the transcription is \textsc{elan}~\citep{elan2024}, that allows for precise time-alignment.
The conversations are segmented into transcription units (TUs) based on prosodic and/or semantic criteria and content is transcribed in accordance with Italian orthographic norms and a set of conventions to represent prosodic and interactional features inspired by Jefferson notation~\citep{lernerglossary2004}: the conversational conventions adopted include symbols to mark, for instance, overlapping speech, pauses, prolongations, non-verbal behaviour e.g., \textit{((applaude))}, `claps' \textit{((ride))}, `laughs', and so on (see Table~\ref{tab:example} for an example from the current experiment and Table~\ref{tab:jefferson} for a complete reference to the set of symbols employed in transcription).
In order to get from source audio to fully queryable transcript, a team of linguists with various areas of expertise are involved, including properly trained and regularly monitored student interns. 
Transcriptions produced by interns are then fully revised by an expert linguist, in order to mitigate differences introduced by transcribers that might tend to over- or under-represent given phenomena, and to cross-check internal consistency.
However, it is important to underline how even revised content cannot be considered ``Gold standard'' in the sense that is usually intended in NLP research. In linguistic research, data are not direct reflections of physical phenomena, but rather representation of linguistic events. Consequently, within this context, transcriptions are not an objective reflection of speech, but rather abstract and interpretative acts shaped by methodological choices and theoretical frameworks. In this sense, it is impossible to escape subjectivity, which is not considered 'noise' to eliminate, but an integral part of the data itself \cite{Lehmann2004DataIL}. 

\begin{table*}[htbp]
\centering
\resizebox{\linewidth}{!}{%
\footnotesize{
\begin{tabular}{p{0.05\linewidth}p{0.05\linewidth}p{0.05\linewidth}p{0.4\linewidth}p{0.4\linewidth}}
\hline
\textbf{speaker} & \textbf{start} & \textbf{end} & \textbf{transcription} &  \textbf{translation} \\ \hline
SP2 & 3.14 & 6.07 & sì (.) non era pesantissima & yes (.) it wasn't too rich\\
SP1 & 6.34 & 7.52 & per niente {[}(ha-){]} & not at all {[}(the-){]}\\
SP2 & 7.22 & 10.09 & {[}uni{]}ca cosa, ho esagerato un po' con l'olio forse però & {[}the on{]}ly thing is, I exaggerated a bit with oil maybe but\\
SP1 & 10.85 & 12.28 & ma nell'impasto? no fuori, & but in the dough? no outside,\\
SP2 & 12.46 & 15.76 & °fuori° (.) nell'im{[}pasto ce ne ho messo il{]} giusto però in padella & °outside° (.) in the d{[}ough I put just{]} enough but in the pan\\
SP3 & 13.39 & 14.36 & {[}nella padella{]} & {[}in the pan{]} \\
SP2 & 16.97 & 28.63 & xxxx ha detto metti: quello là: l'olio da friggere così vengono meglio ho detto okay (.) e ho fatto però praticamente ho f- (pure) °ho messo:° tutto il fondo della padella era pieno di olio & xxxx said pu:t that there: the frying oil so they cook better I said okay (.) and I did it but in practice I d- (also) °I pu:t° the whole bottom of the pan was full of oil\\
\hline
\end{tabular}%
}
}
\caption{Extract of a transcription of a free conversation from KPS021, KIPasti module \citep{kipastilr}. Seven transcription units are shown, with their respective time boundaries, uttered by three different speakers. Text is enriched with Jefferson-style notation.}
\label{tab:example}
\end{table*}

\begin{table}[h]
\centering
\footnotesize{
\begin{tabular}{p{0.3\linewidth} p{0.6\linewidth}}
\hline
\textbf{Symbol} & \textbf{Meaning} \\
\hline
\verb|word,| & Weakly rising intonation\\
\verb|word?| & Rising intonation\\
\verb|word.| & Falling intonation \\
\verb|wo:rd| & Prolonged sound \\
\verb|(.)| & Short pause \\
\verb|word=word| & Prosodically linked units\\ 
\verb|°word°| & Lower volume\\
\verb|WORD| & Higher volume\\
\verb|wor-| & Interrupted word \\
\verb|>word<| & Faster speech\\
\verb|<word>| & Slower speech \\
\verb|[word]| & Overlapping speech\\
\verb|(word)| & Uncertain transcription (transcriber’s guess)  \\
\verb|xxx| & Unintelligible sequence (one x per syllable)  \\
\verb|((laughs))| & Non-verbal behavior  \\
\hline
\end{tabular}
}
\caption{Transcription conventions}
\label{tab:jefferson}
\end{table}

\section{Literature Review}
\label{sec:literature}

When dealing with conversational data, transcription extends beyond speech-to-text conversion: speech is spontaneous and interactionally complex and it can also be affected by audio quality, overlapping turns and linguistic variation~\cite{Ghyselen2020,Liesenfeld2023}. Elements such as backchannels, interjections, filled pauses, and paralinguistic cues~\cite{Umair2022,Zayats19} pose significant transcription challenges. Even among human transcribers disagreements mainly involve function words and backchannels, while content words are less prone to error~\cite{Zayats19}.
ASR systems, however, are trained primarily on standard, monologic speech and tend to ignore conversational features, treating them as noise~\cite{Lopez2022,Yamasaki2023}. As a result, although they perform well in controlled settings, their performance decreases in spontaneous and real-world conversational contexts~\cite{Cumbal2024,Gaur2016}.As far as the evaluation is concerned, transcription accuracy is commonly evaluated using the Word Error Rate (WER, \citep{klakow_testing_2002}), where a lower WER indicates higher transcription quality. \citet{Gaur2016} suggest that a WER above 30\% marks the point where manual transcription becomes more efficient than ASR-assisted transcription. However, it has been criticized for conversational data: \citet{Liesenfeld2023} argue that WER oversimplifies performance and ignores key discourse features such as disfluencies or overlaps, while \citet{Gorisch2024} note that WER tends to exaggerate omissions, especially of hesitation markers and backchannels. Despite these limitations, WER remains the standard metric for ASR evaluation, and, to out knowledge, no widely accepted alternatives designed for conversational contexts currently exist. For all these reasons, the literature provides limited support for ASR-assisted transcription workflows. The most comprehensive study to date~\citep{Gorisch2024} reports that correcting ASR output requires comparable time and effort to manual transcription. Similar findings are reported by \citet{Ghyselen2020}, while \citet{Liesenfeld2023} emphasize that current ASR systems prioritize lexical content over temporal and interactional features. However, these studies often rely on datasets with dialectal variation or suboptimal recording conditions, which may limit the generalizability of their conclusions. Within the KIParla context, an additional methodological difference is relevant: in the design adopted by \citet{Gorisch2024}, manual and ASR-assisted workflows are not applied to the same audio segments, preventing direct one-to-one comparison. Their results nevertheless highlight that transcription time is more strongly influenced by transcriber identity and expertise than by the transcription method itself. This factor is particularly relevant for KIParla, where the corpus is mainly produced by student interns with limited transcription experience and high turnover. Consequently, developing a reliable methodology that can support novice transcribers may be especially beneficial.

\section{Data, Experimental Setup and Research Questions}
\label{sec:experiment}

The aim of our experiment is to explore the possibility of introducing an ASR transcription step in the current KIParla workflow, as a preliminary step: transcribers would be therefore asked to correct the ASR output instead of starting the transcription task from scratch. To carry out the experiment, 11 transcribers (4 ``expert'' transcribers and 7 ``novices'', based on their transcription experience\footnote{Interns who had already transcribed at least 45 minutes of speech were considered experts, see Table~\ref{tab:participants-experience} in Appendix.}) were recruited among interns on a voluntary basis. 

We selected seven 10-minute extracts from available conversations, showcasing different features: two extract were selected from \textit{free conversation} (see Table~\ref{tab:example}), two extracts of \textit{semi-structured interviews} (see Table \ref{tab:posto_cuore}) and three \textit{interviews with L2 material} (see Table \ref{tab:bangla}). Full statistics about the extracts are reported in Table~\ref{tab:extracts}. As the examples illustrate, the conversations differ in several respects, each of which may influence the outcome of the experiment in distinct ways.
Kitchen‑table conversations tend to show shorter turns, more frequent overlaps, and a higher rate of speaker changes. In contrast, semi‑structured interviews are predominantly monologic. Finally, interview SBIB003 is characterised by reduced fluency, a greater number of self‑repairs, and less predictable lexical choices.

\begin{table}[t]
\centering
\small
\setlength{\tabcolsep}{4pt}
\begin{tabularx}{\columnwidth}{
    >{\raggedright\arraybackslash}X
    c
    c
    c
}
\hline
\textbf{Conversation type} & \textbf{Extract} & \textbf{Tokens/min} & \textbf{Types/min} \\
\hline
\multirow{2}{=}{Free conversation}
  & A & 183.23 & 64.39 \\
  & B & 199.31 & 78.08 \\\hline
\multirow{2}{=}{Semi-structured interview}
  & A & 172.30 & 52.85 \\
  & B & 170.80 & 51.04 \\\hline
\multirow{2}{=}{Interviews with L2 material}
  & A & 199.45 & 61.49 \\
  & B & 180.61 & 70.15 \\
\hline
\end{tabularx}
\caption{Tokens and Types per minute in ``Gold'' transcriptions for the 6 considered 10-minute extracts.}
\label{tab:extracts}
\end{table}

\begin{table*}[htbp]
\centering
\resizebox{\linewidth}{!}{%
\begin{tabular}{p{0.08\linewidth}p{0.05\linewidth}p{0.07\linewidth}p{0.4\linewidth}p{0.4\linewidth}}
\hline
\textbf{speaker} & \textbf{start} & \textbf{end} & \textbf{transcription} & \textbf{translation} \\ \hline
SP1 & 684.82 & 687.56 & il tuo posto del cuore di bologna & the place of your heart in Bologna \\
SP1 & 688.01 & 689.78 & non so anche solo un bar & I don't know maybe just a cafeteria \\
SP2 & 689.80 & 690.29 & {[}mh{]} & [mhm] \\
SP1 & 689.83 & 691.95 & {[}a cui se{]}i particolarmente affezionata, & [that you]'re particularly fond of \\
SP2 & 692.63 & 693.13 & mh & mhm \\
SP2 & 695.10 & 697.55 & e:h forse: mh:h & e:h maybe uh:m \\
SP2 & 698.37 & 702.64 & mangiare un gelato seduti sul gradino di piazza santo stefano ((ride)) & having an ice-cream while sitting on a step in Piazza Santo Stefano ((laughs)) \\
SP1 & 701.80 & 702.39 & a:h & a:h \\
SP1 & 702.64 & 703.74 & lì è bellis{[}simo{]} & it's really beautiful th[ere] \\
SP2 & 703.21 & 705.34 & {[}perchè lì{]} piazza santo stefano mi piace molto & [because I] really like Piazza Santo Stefano \\
SP2 & 705.36 & 707.39 & poi c'ha di fianco la mia gelateria preferita, & it's also near my favourite ice-cream place \\
SP2 & 707.64 & 709.97 & e quindi spesso d'estate col mio ragazzo andiamo & so we often go in the summer with my boyfriend \\
SP2 & 710.02 & 714.66 & andiamo lì prendiamo il gelato e ci sediamo: in piazz{[}a: santo s{]}tefano >°quindi forse quello {[}sì°{]}< & we go there have an ice-cream and sit do:wn in Piazza: Santo Stefano >°so maybe that {[}yes°{]}< \\
\hline
\end{tabular}%
}
\caption{Extract of a transcription of a semi-structured interview from PBA001, ParlaBO module \citep{parlabolr}). Text is enriched with Jefferson-style notation.}
\label{tab:posto_cuore}
\end{table*}

The experiment consisted of two transcription sessions. During Phase 1 (``Manual''), participants transcribed an extract from scratch using \textsc{elan}, following the usual pipeline. During Phase 2 (``ASR-assisted'') they received an ASR-generated textual transcription, obtained via \texttt{openai-whisper} large model~\cite{radford2023robust}: they manually assigned speaker turns, then revised the ASR output by importing \texttt{.srt} files, one per speaker, into \textsc{elan}, where they also added Jefferson notation.\footnote{The only exception with respect to the usual pipeline was the use of $=$, which was excluded both to simplify the task and because it proved particularly difficult to apply by interns.}
The extracts were assigned adopting a schema that guaranteed to obtain at least two versions, one manual and one ASR-assisted, for each extract. It was also ensured that the same participant did not work on the same extract twice.
Each transcription session lasted 120 minutes. Participants were asked to record their transcription progress every 30 minutes, indicating the number of seconds transcribed, as shown in Table~\ref{tab:transcribed-times}. For each of the selected segments, we also collected an independently produced Gold standard transcription, which is the one revised by the expert linguist and currently available for consultation\footnote{For ``L2-interviews'', which are not yet available for consultation, we autonomously produced a Gold version for comparison.}.

Prior to analysis, all transcriptions were cleaned and standardized to ensure consistency. 
This included removing characters such as spurious newlines, tabs, double spaces, non-Jefferson symbols and short pauses at the beginning or end of a transcription unit (TU). 
Orthographic corrections were then applied to restore standard spellings (e.g., \textit{pò} > \textit{po'},  \textit{perchè} > \textit{perché}, etc.). Bracket Jefferson symbols  
were verified for proper balance and formatting across types (e.g., \texttt{°°}, \texttt{[]}, \texttt{(.)}, \texttt{()},\texttt{<>},\texttt{><}).
Spacing consistency around brackets and punctuation was normalised and numbers were converted from digits into words. After pre-processing, transcriptions were normalised, tokenised, and Jefferson notation was transformed into token-level features. 

\begin{table*}[htbp]
\centering
\resizebox{\linewidth}{!}{%
\begin{tabular}{p{0.08\linewidth}p{0.05\linewidth}p{0.05\linewidth}p{0.4\linewidth}p{0.4\linewidth}}
\hline
\textbf{speaker} & \textbf{start} & \textbf{end} & \textbf{transcription} & \textbf{translation} \\ \hline
SP1 & 830.44 & 832.43 & parla- m:h che lingua parlate? & you spea- uh:m which language do you speak? \\
SP2 & 832.79 & 835.93 & [bangla] sì è la lingua:: bangla che parlan[o] & [bangla] yes it's the bangla:: language they spe[ak] \\
SP1 & 832.79 & 833.33 & [quale li:-] & [which la:-] \\
SP1 & 835.86 & 836.34 & [e:] & [a:nd] \\
SP2 & 836.17 & 839.12 & [p]erò eh tutti quanti lo sanno anche italiano °tranquilla[mente°] & [b]ut eh everybody has no problem with Italian °as [well]° \\
SP1 & 838.72 & 843.35 & [okay] e lo mi- mischiate un po' [con l'italiano] quando parlate fra di [voi o] & [okay] and you mi- mix it [with Italian] a bit when you speak among each [other or] \\
SP2 & 840.68 & 841.36 & [sì sì] & [yes sure] \\
 & & & {[}\dots{]} & {[}\dots{]} \\
SP2 & 870.55 & 879.56 & no esempio è quando noi parliamo qualche argomenti di proprio cioè proprio ital- c'è un posto qui allora parliamo come: & no an example is when we speak some topics of exactly I mean exactly ital- there is a place here so we speak like: \\
SP2 & 879.81 & 881.86 & come italiano cioè proprio le cose & like Italian I mean exactly the things \\
SP2 & 881.91 & 890.27 & però eh quando noi parliamo la roba di per la famiglia (.) allora vabbè utilizziamo il lingua bangla per per la famiglia assolutamente & but eh when we speak stuff of for the family (.) so then we use bangla language for for the family absolutely \\
\hline
\end{tabular}%
}
\caption{Extract of a transcription of an L2 interview from SBIB003, StraParlaBO module (see \citealt{Stra-ParlaBO,mauri2025il} and \citet{straparlabolr}). Text is enriched with Jefferson-style notation.}
\label{tab:bangla}
\end{table*}

\begin{table*}[htbp]
\centering
\resizebox{\textwidth}{!}{%
\begin{tabular}{l|lcccc|lcccc}
\hline
 &  \multicolumn{5}{c|}{\textbf{Manual Phase}} & \multicolumn{5}{c}{\textbf{ASR-assisted Phase}} \\
\textbf{Transcriber} & \textbf{conversation} & \textbf{30 min} & \textbf{60 min} & \textbf{90 min} & \textbf{120 min} & \textbf{conversation} & \textbf{30 min} & \textbf{60 min} & \textbf{90 min} & \textbf{Final} \\
\hline
A &  L2-interview$_2$ & 60 & 129 & 193 & 232 & L2-interview$_1$ & 74 & 110 & 153 & 195 \\
E &  interview$_1$ & 117 & 268 & 378 & 503 & free-conversation$_1$ & 113 & 257 & 360 & 465 \\
I &  L2-interview$_1$ & 69 & 138 & 229 & 254 & interview$_1$ & 71 & 133 & 206 & 290 \\
O &  free-conversation$_1$ & 120 & 220 & 396 & 406 & L2-interview$_2$ & 90 & 180 & 254 & 304 \\
\hline
L & L2-interview$_1$ & 51 & 118 & 224 & 280 & interview$_1$ & 41  & 79 & 130 & 175 \\
M & interview$_1$ & 48 & 90 & 149 & 212 & free-conversation$_1$ & 67 & 162 & 239 & 284 \\
N & interview$_2$ & 70 & 144 & 213 & 242 & L2-interview$_3$ & 64 & 117 & 180 & 187 \\
R & free-conversation$_1$ & 40 & 72 & 135 & 185 & L2-interview$_2$ & 64 & 104 & 173 & 191 \\
S & free-conversation$_2$ & 180 & 300 & 410 & 491 & interview$_2$ & 49 & 110 & 155 & 197 \\
U & L2-interview$_3$ & 82 & 147 & 236 & 265 & free-conversation$_2$ & 42 & 77 & 110 & 145 \\
V & L2-interview$_2$ & 53 & 73 & 138 & 185 & L2-interview$_1$ & 69 & 149 & 217 & 251 \\
\hline
\end{tabular}%
}
\caption{Audio extracts assigned to both expert (top pane of the table) and novice transcribers (bottom pane) in the two phases of the experiment, along with transcription times comparison. In the ASR-assisted phase only three measures were collected, as the first part of the session was used to pre-process the Whisper output. Measures are expressed in seconds. See also Figures~\ref{fig:transcriptiontimes} and \ref{fig:transcriptiontimes2} in Appendix.}
\label{tab:transcribed-times}
\end{table*}

Given this setting, we investigate two main research question: (RQ1) Can automatic transcription help speed up the corpus creation process, without compromising quality? and (RQ2) Can automatic transcription help us better monitor the internal consistency of the resource?

\section{Results}
\label{sec:results}
\subsection{Descriptive Statistics and Comparison with ``Gold''}
\label{sec:stats}
As a first step to assess similarity across transcriptions, a comprehensive set of statistics was generated, including general metrics (e.g., total number of TUs, average TU length), several per-minute statistics (covering TU duration, counts of linguistic tokens and non-verbal-behaviors, number of short pauses, errors and unknown tokens), and average measures (e.g., average number of tokens per TU, average TU duration).  
Finally, each transcription was linked to annotator metadata, particularly their transcription experience.

To evaluate the similarity of ``Manual'' and ``ASR-assisted'' annotations to the ``Gold'', delta values were calculated as the difference in the computed value for the Gold transcript and for each of the transcripts produced by participants, namely, $\Delta_M$ for Manual transcripts, and $\Delta_A$ for ASR-assisted ones. 
These deltas were averaged across extracts for the three data types (\textit{interviews}, \textit{free-conversations}, \textit{L2-interviews}) for both expert and novice transcribers. 
To ensure consistency and avoid missing data, all deltas were computed over the first two minutes of transcription (that all participants reached) and rounded to two decimal places. 

Deltas closer to zero indicate greater similarity to the gold standard. When $|\Delta_A|<|\Delta_M|$, the ASR-assisted transcription is more similar to the gold; when $|\Delta_M|<|\Delta_A|$, the Manual approach yields a more faithful transcription. Opposing signs suggest that the two methods behave differently in relation to the Gold: a positive $\Delta$ indicates that intervention of the reviewer would \textit{add} something (e.g., provide a more fine-grained segmentation into TUs, add tokens etc.,), while a negative $\Delta$ signals the opposite. 
Table~\ref{tab:deltas} shows delta values for two metrics: number of tokens per minute and number of transcription units (TUs) per minute. Concerning the former, the interview module shows relatively small deltas for ASR-assisted transcriptions (e.g., –1 at minute 1), suggesting closer alignment with the Gold output. Manual transcriptions diverge more notably (e.g., –16 at minute 2), indicating underproduction. In free conversation, both workflows tend to overproduce, but ASR-assisted shows larger deltas (e.g., +41 at minute 2), pointing to greater deviation from the Gold. In L2-interviews, both workflows show inconsistent alignment: Manual deltas are positive and stable (+16), while ASR-assisted values fluctuate, ending with a drop at minute 2 (–22). Turning to the number of TUs per minute, ASR-assisted transcriptions in the interview show the highest deltas (e.g., +28.17 at minute 2), reflecting finer segmentation than the Gold, while Manual also oversegments but less dramatically. In free conversation, deltas are generally small, with Manual slightly closer to the reference by minute 2. In the L2-interview, both workflows begin with high TU deltas but decrease over time, with ASR-assisted falling more sharply, suggesting an increase in undersegmentation.

\begin{table}[htbp]
\centering
\resizebox{\columnwidth}{!}{%
\begin{tabular}{l|l|rrr|rrr}
\hline
 &  & \multicolumn{3}{l}{tokens per minute} & \multicolumn{3}{l}{TUs per minute} \\ \hline
\textbf{Module} & $\Delta$ & \textbf{0} & \textbf{1} & \textbf{2} &  \textbf{0} & \textbf{1} & \textbf{2}  \\ \hline
interview & $\Delta_M$ & $\downarrow$5 & $\downarrow$8 & $\downarrow$16 & $\uparrow$4.67 & $\uparrow$11.33 & $\uparrow$20.17 \\
interview & $\Delta_A$ & $\uparrow$2 & $\downarrow$1 & $\downarrow$8 & $\uparrow$7.67 & $\uparrow$18.33 & $\uparrow$28.17 \\
free-conversation & $\Delta_M$ & $\uparrow$6 & $\uparrow$15 & $\uparrow$29 & $\uparrow$6.83 & $\uparrow$11.00 & $\uparrow$14.33 \\
free-conversation & $\Delta_A$ & $\uparrow$4 & $\uparrow$22 & $\uparrow$41 & $\uparrow$1.83 & $\uparrow$2.00 & $\uparrow$0.33 \\
L2-interview & $\Delta_M$ & $\uparrow$2 & $\uparrow$15 & $\uparrow$16 & $\downarrow$7.47 & $\downarrow$16.67 & $\downarrow$20.53 \\
L2-interview & $\Delta_A$ & $\downarrow$11 & $\uparrow$8 & $\downarrow$22 & $\uparrow$4.53 & $\uparrow$7.73 & $\uparrow$2.87 \\ \hline
\end{tabular}%
}
\caption{Deltas in tokens per minute and number of transcription units per minute across modules and transcription methods (Gold vs. Manual/ASR-assisted) at time intervals 0–2 minutes. Positive values are identifies by a $\uparrow$ symbol while negative values by a $\downarrow$ one.}
\label{tab:deltas}
\end{table}

\subsection{Word-level Alignment and Word Error Rate}

Transcriptions of the same recording were aligned against their ``Gold'' counterpart through a word-level alignment pipeline. The process relies on the Needleman-Wunsch algorithm~\cite{NEEDLEMAN1970443} to identify matches, insertions, deletions and substitutions needed to align the two token sequences. To guarantee a valid comparison, the alignment was constrained to the portion of the transcript shared by both files, based on their temporal overlap. 
This alignment process was carried out to facilitate the following computation of the Word Error Rate \citep{klakow_testing_2002}, which is defined as ${WER}=\frac{S + D + I}{N} = \frac{S + D + I}{S + D + C}$ with $S$ (substitutions), $D$ (deletions), and $I$ (insertions) representing errors, $C$ the number of correct words, and $N$ the total number of words in the reference (i.e., Gold) transcript. The output consists of aligned token sequences where mismatches are explicitly marked with gap symbols, allowing for systematic comparison across annotation layers. Despite the limitations discussed in \ref{sec:literature}, we introduce it as an exploratory metric.

Table~\ref{tab:wer} shows how the shift from a Manual pipeline to an ASR-assisted one improves the faithfulness of transcription for 7 participants out of 11. While no clear trend emerges, ASR-assisted transcriptions tend to yield lower WER values than Manual ones, especially among non-expert annotators. For example, transcribers S, U, V, none of
whom had prior transcription experience, produced manual transcriptions with notably high WERs,
respectively, 39.37\%, 29.3\% and 32.67\%. With the support of ASR, these values dropped
significantly to 29.45\%, 22.72\% and 22.26\%.
Conversely, expert annotators show more heterogeneous results, suggesting that
many factors might interact to shape the effectiveness of ASR implementation, such as individual transcription strategies.

\begin{table*}[]
\begin{tabular}{ll|ll|ll|c}
\hline
 &  & \multicolumn{2}{c}{\textbf{Manual}} & \multicolumn{2}{c}{\textbf{ASR-assisted}} &  \\ \hline
\textbf{Transcriber} & \textbf{Expert} & \textbf{Data} & \textbf{WER} & \textbf{Data} & \textbf{WER} & $\text{WER}_M - \text{WER}_A$ \\ \hline
A & Yes & L2-interview & 26,35 & L2-interview & 15,14 & 11,21 \\
E & Yes & interview & 15,92 & free-conversation & 17,99 & \textbf{-2,07} \\
I & Yes & L2-interview & 20,09 & interview & 24,46 & \textbf{-4,37} \\
L & No & L2-interview & 25,04 & interview & 22,74 & 2,3 \\
M & No & interview & 20,9 & free-conversation & 29,77 & \textbf{-8,87} \\
N & No & interview & 22,36 & L2-interview & 16,37 & 5,99 \\
O & Yes & free-conversation & 31,39 & L2-interview & 19,49 & 11,9 \\
R & No & free-conversation & 18,29 & L2-interview & 22,87 & \textbf{-4,58} \\
S & No & L2-interview & 39,37 & interview & 29,45 & 9,92 \\
U & No & L2-interview & 29,3 & free-conversation & 22,72 & 6,58 \\
V & No & free-conversation & 32,67 & L2-interview & 22,36 & 10,31 \\
\hline
\end{tabular}
\caption{WER Analysis by Transcriber Expertise, Transcription Method and Deltas}
\label{tab:wer}
\end{table*}

\subsection{Overlap Classification}
\label{sec:preliminary}

As far as overlap errors are concerned, each inconsistency reported by the automatic pre-processing script was manually checked and rated according to the following criteria: \textit{severe}, when there was a temporal overlap, but it had not been annotated, or the opposite; \textit{mildly-severe}, when the conditions above applied, but the temporal overlap was minor (around 0.1 seconds); when one longer TU overlapped with two shorter ones, but the transcriber inserted only one notation on the longer TU; \textit{non-severe}, in case of repeated consecutive parenthesis ([[) or unclosed parenthesis.
Cases of overlap with non-verbal behaviour (such as ((\textit{ride})), 'laughs') were ignored.
The analysis of the distribution of overlap errors did not take into account the type of conversation transcribed, since, in this respect, the data are not balanced among novice and expert transcribers. 
By counting the raw number of overlap errors in transcriptions made by each group, we notice that the use of ASR systems seems to benefit more expert transcribers, for which the total number of errors drops from 26 to 15. For novices, instead, the total number remains stable (61 in the manual phase, 60 in the ASR-assisted one).

\begin{figure}
    \centering
    \includegraphics[width=0.7\linewidth]{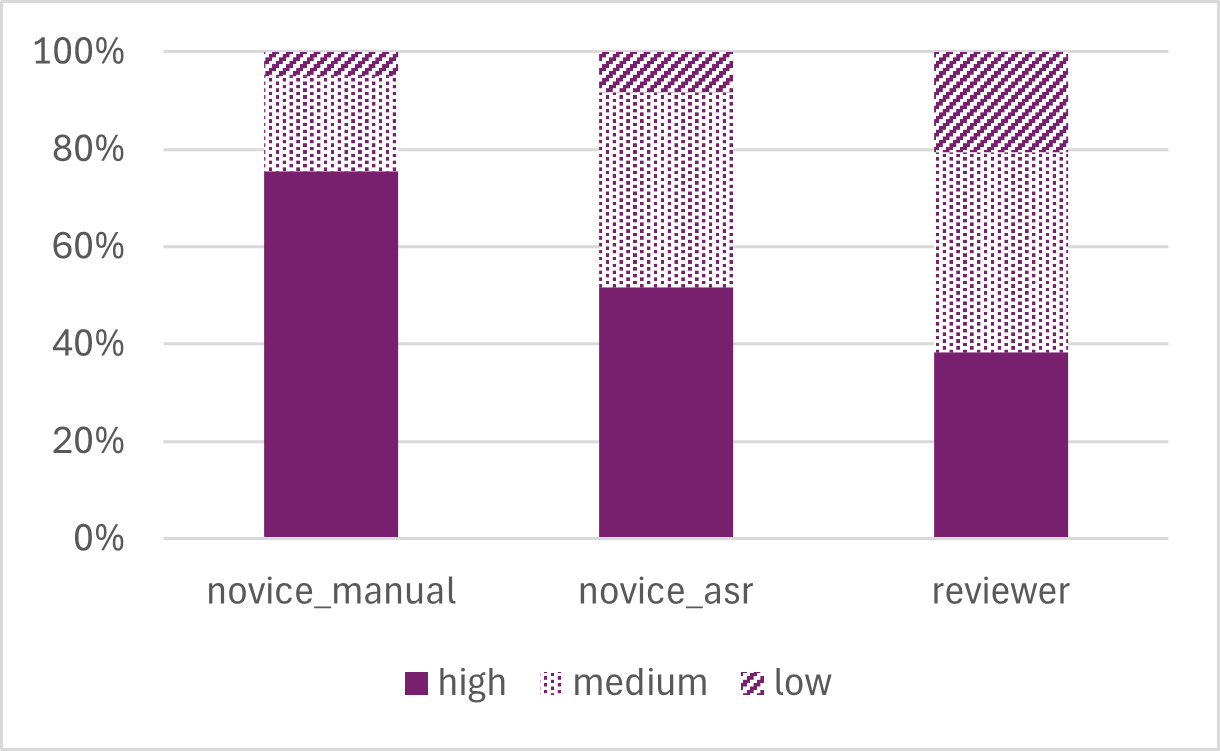}
    \caption{Percentage of overlap errors, novices' transcriptions}
    \label{fig:overlap_novice}
\end{figure}

\begin{figure}
    \centering
    \includegraphics[width=0.7\linewidth]{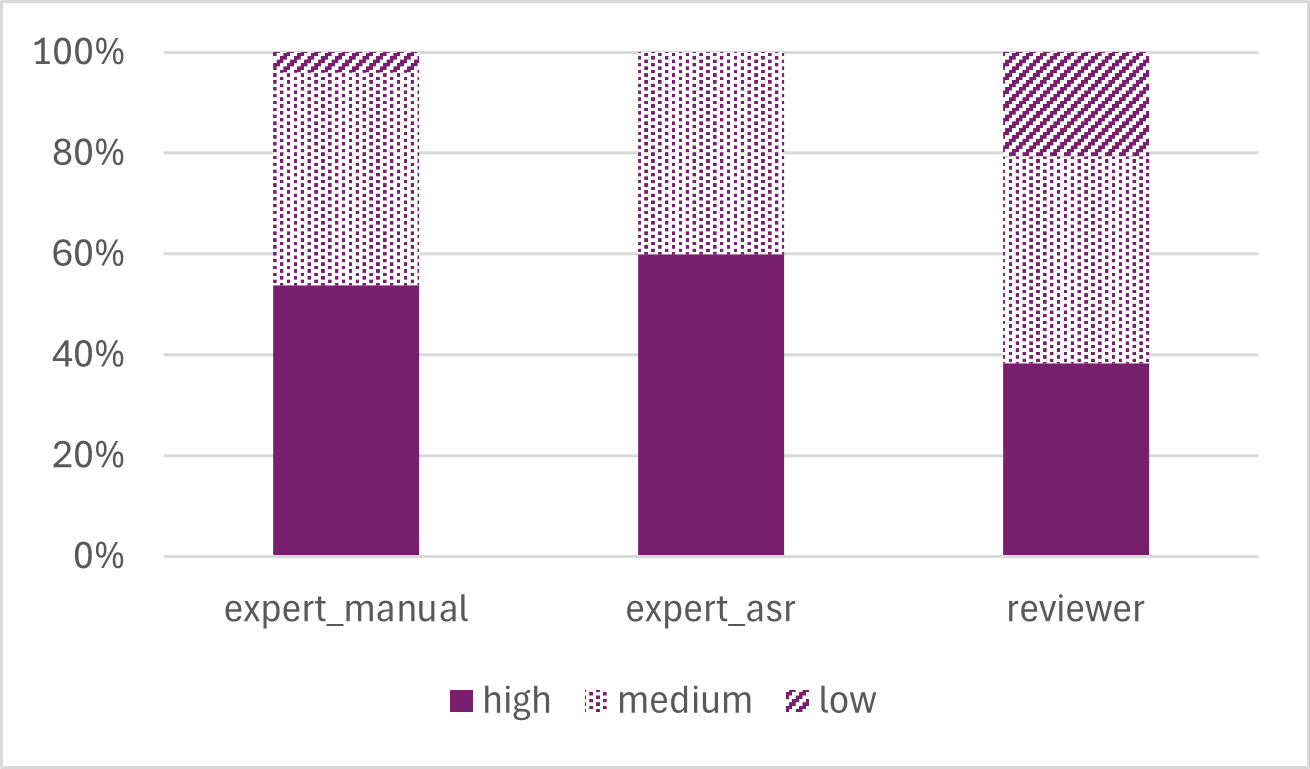}
    \caption{Percentage of overlap errors, experts' transcriptions}
    \label{fig:overlap_experts}
\end{figure}

The manual revision allows to make a closer inspection of the degree of severity of the overlap errors made by each group (three levels: severe, mild and non-severe); Table \ref{tab:overlap_error} and Figures \ref{fig:overlap_novice} and \ref{fig:overlap_experts} show the results of this comparison and also include the percentage of each kind of error made by the expert reviewer, which is reported only to be considered as a baseline.

\begin{table}[htbp]
\centering
\begin{tabular}{l l l l}
\hline
 & \textbf{severe} & \textbf{mild} & \textbf{non-severe} \\
 \hline
\textbf{novice\_manual} & 75\% & 20\% & 5\% \\
\textbf{novice\_ASR} & 52\% & 40\% & 8\%\\ \hline
\textbf{expert\_manual} & 54\% & 42\% & 4\% \\
\textbf{expert\_ASR} & 60\% & 40\% & 0\% \\ \hline
\textbf{reviewer} & 38\% & 41\% & 21\% \\
\hline
\end{tabular}
\caption{Percentage of overlap errors according to type and expertise}
\label{tab:overlap_error}
\end{table}

Generally speaking, the expert reviewer makes a more balanced amount of each kind of error, in comparison to both groups and with both pipelines. 
Considering experts alone, this analysis does not add much to the picture drawn before, since the type of error they make is not affected by the use of ASR to a great extent, especially when compared to the group of novices: the latter, as a matter of fact, when using ASR, make significantly less severe mistakes compared to manual transcription (75\% vs 52\%). With the help of ASR, the behaviour of novice transcribers becomes more similar to that of the expert reviewer in qualitative terms. Overall, this seems to suggest that using an ASR system impacts both expert transcribers, in terms of total number of errors, and novice transcribers, in terms of error severity. 

\subsection{Statistical Modeling}

In order to evaluate the influence of factors such as transcriber's expertise, data type and transcription settings on the produced output, we fitted a series of Generalized Additive Mixed Models (GAMMs), as they allow us to account for both fixed effects (i.e., experimental phase, expertise and data type) and nonlinear trends over time using a smooth function of minutes, which represents the time elapsed since the experiment began. Additionally, by including a random intercept for each transcriber, the model accounts for individual differences in transcription behavior, providing more reliable estimates of the effects of interest.
The models were fitted with a Gaussian distribution, selected via Restricted Maximum Likelihood (REML) over Poisson and Gamma alternatives due to superior fit. Our formula ($X \sim \text{phase} + \text{expert} + \text{data} + s(\text{minutes}) + (1|\text{transcriber})$)
was employed for a number of dependent variables ($X$), namely (i) seconds of audio transcribed over a 30 minutes time span (RQ1), (ii) per minute statistics (see Section~\ref{sec:stats}) measuring TU length (in terms of tokens and seconds), number of transcribed tokens (including and excluding non-verbal behaviours), some Jefferson-derived measures (number of annotated intonation patterns and prolongations) and number of annotated overlaps (RQ2), and (iii) per minute $\Delta$ values of the same aforementioned measures (RQ2).The rationale was to assess whether the introduction of ASR in the pipeline would influence transcription speed, while also checking if specific features of the ASR output (i.e., lack of overlapping speech in transcription, normalization of disfluencies, and tendency to segment in larger, meaning-based chunks) would influence the transcribers' work.
Due to the relatively small sample size, no interaction factors were considered.

When fitted to seconds of transcribed audio (RQ1), results from the GAMM indicate that both expertise and experimental phase significantly affected performance. Experts transcribed faster than novices ($\beta = -26.38, p = 0.002$), and the ASR-assisted pipeline was associated with a significant increase in transcription activity compared to the Manual pipeline ($\beta = 27.11, p < 0.001$). The type of data also played a role: the L2-interview was associated with a reduction in transcription delta relative to the reference category ($p = 0.012$).
However, the random intercepts for transcribers were significant ($edf = 7.10, p < .001$), indicating considerable individual variability. The model accounted for $67.2\%$ of the deviance, with an adjusted $R^2$ of $0.595$\footnote{Residual diagnostics indicated no major violations of model assumptions: residuals were approximately normally distributed and homoscedastic, with no evidence of nonlinearity or high-leverage outliers, supporting the adequacy of the model fit. Full results are reported in Appendix, Table~\ref{tab:models}.}.
Introducing ASR in the pipeline is associated with significantly higher transcription activity, with an effect that is comparable to that associated with transcriber expertise (see also Figure~\ref{fig:effects} in Appendix).

As far as RQ2 is concerned, among the 20 fitted models, 11 exhibited $R^2 > 0.45$; the analysis will focus solely on these.
The model fitting average TU length exhibited the strongest fit ($R^2 = 0.822$). 
Smooth terms for transcribers were significant in almost all models\footnote{$p < .05$}, confirming considerable individual differences that were appropriately modeled with random effects. 
The models consistently found expertise to be non-significant across all outcome variables ($p > 0.1$ in each case), except for the one about the number of overlaps, which was consistent with what was described in Section~\ref{sec:preliminary}. This suggests that, at least within the scope of the current dataset and experimental setup, being an expert or novice does not systematically affect transcription performance in terms of structure or output. This result is notable and somewhat unexpected given assumptions about expertise yielding greater efficiency or higher-quality output. However, it is possible that the metric used (a binary classification) may not fully capture the nuanced skills of transcribers. Additionally, individual variability captured in the random effects could be masking more subtle group differences.
The experimental phase emerged as a significant predictor in multiple models. Specifically, the manual phase was associated with an increase in the number of transcription units ($p = .032$) and a decrease in average transcription unit length ($p < .001$). This pattern implies that during manual transcription, transcribers produced more but shorter units, perhaps indicating a shift toward finer-grained segmentation. Interestingly, phase had no significant effect on token count, overall duration of TUs, or prolongations. These findings suggest that phase manipulation affected how transcribers structured the data rather than the overall quantity of their output.
The data source had substantial and significant effects on most transcription measures. Compared to a baseline of free conversation, both interview and L2-interview conditions affected the output. In terms of the number of TUs, interviews led to significantly fewer units ($p = .035$), while both the interview ($p = .008$) and the L2-interview ($p = .017$) conditions resulted in significantly fewer tokens. Additionally, both data types increased the average TU length, with highly significant p-values ($p < .001$ and $p = .013$, respectively). 
Together, these findings underscore the complex interplay of task design and data characteristics on transcription behavior. While transcriber expertise showed no significant effects, both the task phase and data type had considerable influence on key outcome variables. The lack of an expertise effect might suggest either a high level of baseline competence among novices or the insensitivity of the chosen metrics to qualitative differences in transcription.
The phase effects point to task design as a modulator of granularity. Manual annotation encourages a more detailed segmentation, possibly because transcribers are consciously reviewing and structuring the material. These patterns are important for designing tools and workflows in speech corpus development, as they indicate that annotation protocols themselves shape the resulting data.

\subsection{Comparison of Transcription Content}

As a final comparison, we aligned (again using \cite{NEEDLEMAN1970443}'s algorithm) \texttt{whisper}'s raw output with the orthographic transcriptions produced in Manual and ASR-assisted phases, and with the Gold output. 
From the aligned texts we identified mismatches, either due to substitutions or insertions/deletions, and manually classified them with respect to the kind of error produced by the model.
We only compared free conversation and semi-structured interviews extracts, and selected a comparable amount of text (ca. 600 tokens) to manually investigate.
A number of errors depend on \textbf{spelling and orthographic norms}: Whisper and annotators make different choices with respect to the conventional transcription of items such as the continuer ``mh'', which is transcribed as ``mmm'' by Whisper. These cases could be easily treated by postprocessing steps to the transcription. One interesting case is that of ``fiordilatte'' (i.e., a kind of \textit{mozzarella}) which Whisper transcribes as a polirematic item ``fior di latte'': both versions are acceptable in standard Italian orthography. Interestingly, transcribers working on revision decided not to change Whisper suggestion. On the other hand, Whisper tends to normalize standard spellings (e.g., it correctly transcribed ``a parte'' as a two elements item, while one of the transcribers in the Manual phase used the emerging non-standard univerbated orthography ``apparte''), while transcribers seem to be more sensitive to subtle phonetic nuances that may be produced by the speakers in the audio (e.g., transcription of ``bah'' vs. ``beh'' filler). 					
As far as \textbf{interruptions} are concerned, we can distinguish cases where the interrupted content is completely missed by the automatic transcription, and cases where instead Whisper proposed the complete version of the token (e.g., ``salare'', \textit{to salt} in free-conversation was consistently corrected as ``sala-'' by transcribers, same as ``cremonini'', \textit{a proper name}, which is corrected in ``cremoni-'', in semi-structured interview). This seems to depend on the presence of a repetition right after: if no repetition is present, Whisper provides complete transcription.	
We also noticed that Whisper tends to retain vowels that transcribers \textbf{elide}. This includes cases such as ``la avevi'' (en. \textit{you had it}) which was consistenly transcribed as ``l'avevi'' by humans, and cases where elision is not signaled by the apostrophe such as ``sono'' (\textit{to be}, 1st person singular or 3rd person plural) for ``son'', ``particolare'' (en. \textit{particular}) for ``particolar'', which is often elided in front of ``modo'' in Italian. This does not happen systematically so further investigation is needed on the audio signal to assess what could be triggering the behavior.
Some difficulties were encountered with \textbf{proper names}: in the case of ``Cremonini'' (i.e., an Italian singer), Whisper seems to not recognize the tokens as a proper name and adds a masculine plural article (i.e., ``i'', en. \textit{the.PL}), interpreting it as a plural lexical item in accordance with its \textit{morphological appearance} due to the final \textit{-ini}. Similarly the name ``via Indipendenza'' (one of the main streets of Bologna) is misrecognized as ``in pendenza'' or ``indipendenti'', both existing lexical items in the Italian repertoire.
Another class of errors seems to be due to similarities in sounds, that were classified as \textbf{content approximation}. These are cases like  ``suo giorn'' for ``soggiorno'' (en. ``his/her day'' for ``stay''), ``ovunque'' for ``comunque'' (en. ``everywhere'' for ``anyway''), and ``incontrato'' for ``incrociato'' (en. ``met'' for ``bumped into''). The number of these cases seems to be consistently higher for Manual transcriptions, meaning that it might be the case that in ASR-assisted phase transcribers may be more prone to accept Whisper's misheard content. Cases where the misunderstanding produces non existent words such as "diabetto" for ``diabete'' (en. ``diabetes''), ``supernano'' for ``adesso prendiamo'' (en. ``super dwarf'' for ``now we take''), or ``friamenti'' for ``friarielli'' (en. ``broccoli rabe'') are consistently corrected by annotators.
A different case is where Whisper gets the right lexeme, but the error lies in the \textbf{grammatical features} of the form (e.g, a singular form is transcribed as plural, feminine as masculine etc). The number of these kind of correction does not seem  to depend on the kind of pipeline, but further ad hoc investigation is required as such errors could greatly impact subsequent annotation.
Finally, whisper hallucinates content by \textbf{adding} or \textbf{skipping} tokens. In both cases, this seems to concern particles more than content words. The average length of missed and added content is in fact shorter than the average length of transcription content (e.g., $3$ characters versus $4.2$ characters per word on average for \textit{free-conversation}), with no differences with respect to phase.

\section{Participants' Feedback}
\label{sec:participantsfeedback}
At the end of the second experimental phase, each of the participants had the opportunity to give generic feedback regarding the ASR-assisted transcription pipeline. Generally speaking, the machine-assisted pipeline was considered more practical than the completely manual one by the majority of participants (7/11), slightly more practical by two and definitely less practical by the remaining two.

\section{Conclusion and Future Work}
\label{sec:conclusion}

This study analyzed the potential integration of ASR into the transcription workflow of the KIParla corpus by directly comparing manual and ASR-assisted outputs on the same audio stretch. 

Results suggest that the ASR-assisted workflow can increase transcription speed without consistently compromising transcription accuracy (RQ1). The statistical modeling indicates that the ASR-assisted phase is associated with significantly higher transcription activity, with an effect comparable to that of transcription expertise. Word-level alignment and WER analysis further show that ASR assistance often benefits novice annotators, reducing error rates and variability in their outputs, while expert transcribers display more heterogeneous results. These findings suggest that semi-automatic transcription may represent a useful strategy to accelerate corpus creation, particularly when the transcription process involves less experienced annotators.
The analyses also show that combining alignment-based metrics, descriptive statistics and statistical modeling provides a systematic framework to monitor transcription behavior across annotators and workflows (RQ2). The results highlight that transcription output is influenced by the pipeline adopted, characteristics of the data and by individual annotation strategies. In particular, the experimental phase (manual/ASR-assisted) affects segmentation behavior, with manual transcription tending to produce a larger number of shorter transcription units, suggesting a more fine-grained segmentation of the speech signal.
Overall, these findings indicate that the impact of ASR integration in corpus creation is not uniform and depends on a complex interaction between workflow design, annotator expertise and data type. However, the current results remain exploratory due to the limited size and balance of the dataset. Since only a small number of extracts were included for each conversation type, observations may reflect properties of individual recordings rather than systematic differences across interactional settings.

Future work will therefore focus on expanding the dataset and refining the experimental design in order to better disentangle the effects of transcription expertise, conversation type and ASR output quality. In addition, further qualitative analysis of transcription behavior and targeted fine-tuning of ASR models on conversational Italian data may help facilitate the integration of automatic tools into the KIParla transcription pipeline.

\section{Acknowledgements}
The research leading to these results has received funding from Project ”DiverSIta-Diversity in spoken Italian” (PI: Caterina Mauri), prot. P2022RFR8T, CUP J53D23017320001, funded by EU in NextGenerationEU plan through the Italian "Bando Prin 2022 - D.D. 1409 del 14-09-2022".
We would like to thank the participants to the experiment and the support from Laboratorio Sperimentale - LILEC for hosting the transcription sessions.

\section{Limitations}
\label{sec:limitations}
The study has several limitations that should be acknowledged. First, the binary distinction between expert and non-expert transcribers was based solely on the amount of audio previously transcribed (ranging from 45 minutes to 3 hours). While this criterion provides a rough estimate, it might not be enough to measure the actual expertise, which may also depend on other factors, such as the overall complexity of the conversation, data type, the number of speakers or the extent of Jefferson annotations required. Additionally, expertise could be better assessed by factoring in transcription speed in relation to quality: a transcriber who is highly accurate but extremely slow, or one who is fast but produces low-quality output, might not meet the criteria for true expertise. Future studies should adopt a more precise approach, considering both the quantity and quality of transcription experience and possibly establishing clearer criteria. 

Second, transcribers did not work on the same type of conversation in both phases. While, on the one
hand, this decision aimed to avoid repetition and reduce familiarity bias, on the other hand, it also
inevitably limited direct comparative analysis, as only one transcriber worked on the same type of
conversation in both conditions (Transcriber A). Future experiments could address this by selecting
audio samples pertaining to the same module and of comparable difficulty, as well as defining
difficulty metrics through more standardized criteria or annotation guidelines.

Finally, the study would have benefited from more-detailed qualitative insights on the participants' feedback regarding the type of pipeline than what is available:  more thorough interviews with transcribers, as a matter of fact, could help understand the cognitive and strategic effort involved in editing ASR output better. This is especially relevant for Jefferson annotations, where individual strategies may influence results in some way. Future research could also explore the role of transcribers’ familiarity and comfort with automated tools, as this may significantly affect the quality of ASR-assisted transcription. Understanding transcriber behaviour and confidence at this level would be helpful in contextualising
the quantitative findings and improve future annotation protocols.

\section{Ethical Considerations}

The audio data used in the experiment are part of the KIParla corpus, whose recordings were collected in compliance with the corpus ethical and legal protocol. Speakers included in the corpus were informed about the aims of data collection and signed an informed consent form authorizing the recording, transcription, storage and research use of the data. The corpus is managed according to applicable data protection regulations, and personal data are handled with appropriate safeguards. The excerpts used in this study were selected exclusively for research purposes within this framework.

Transcribers involved in the experiment were recruited among student interns and collaborators already participating in corpus-related activities. Their participation in the experiment was voluntary and did not affect their curricular evaluation or internship status. All participants were informed about the purpose of the study and the nature of the tasks involved.

With regard to data processing, the ASR system used in this study was \texttt{openai-whisper}, run locally on institutional devices. No audio data were uploaded to external platforms or processed through third-party online services. This ensured that the speech data remained under the direct control of the research team throughout the transcription workflow.

More generally, the study was designed to evaluate transcription workflows rather than to assess individual participants. Results are therefore reported in anonymized form and are discussed only in aggregate or pseudonymized terms.

\nocite{*}
\section{Bibliographical References}\label{sec:reference}

\bibliographystyle{lrec2026-natbib}
\bibliography{bibliography}


\appendix

\section{Appendix}

\begin{table}[htbp]
\centering
\begin{tabular}{ll}
\hline
\textbf{Participant} & \textbf{Transcription experience}  \\ \hline
O           & 1:06:23                   \\
E           & 2:53:32                   \\
A           & 1:00:51                   \\
I           & 0:46:48                   \\
R,M         & few minutes               \\
L,V,N,U,S   & only trial transcriptions \\ \hline
\end{tabular}
\caption{Transcription experience of the eleven participants of our experiment, expressed in hour:minutes:seconds format.}
\label{tab:participants-experience}
\end{table}

\begin{figure}[h]
    \centering
    \includegraphics[width=0.9\linewidth]{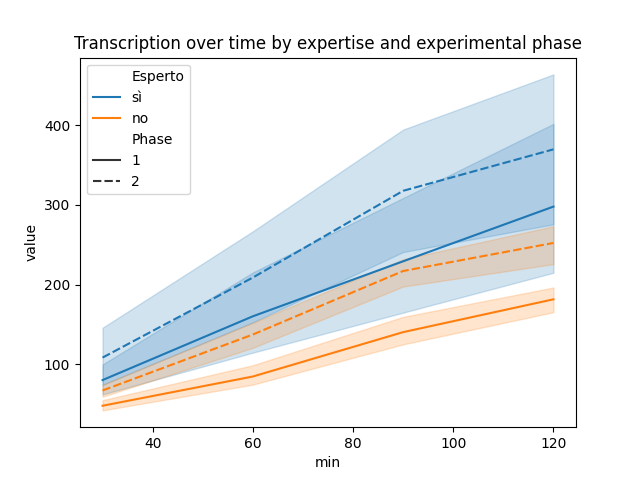}
    \caption{Transcribed minutes by experts (in blue) and novices (in orange) over time during phases 1 (continuous lines) and 2 (dashed lines)}
    \label{fig:transcriptiontimes}
\end{figure}

\begin{figure}[h]
    \centering
    \includegraphics[width=0.9\linewidth]{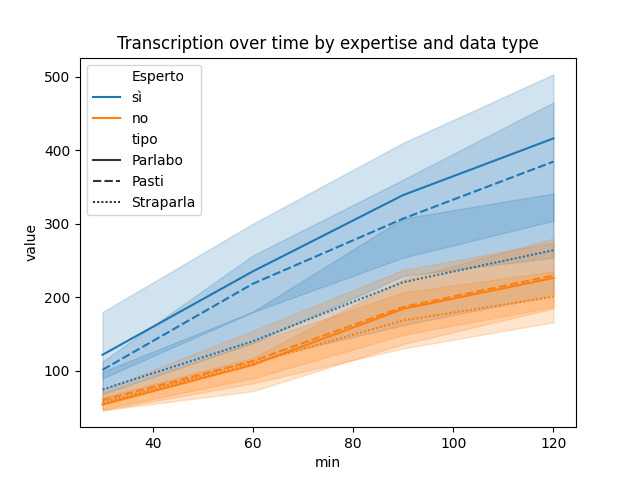}
    \caption{Transcribed minutes by experts (in blue) and novices (in orange) over time for the three different data sources.}
    \label{fig:transcriptiontimes2}
\end{figure}

\begin{figure*}
   \centering
\includegraphics[width=0.9\linewidth]{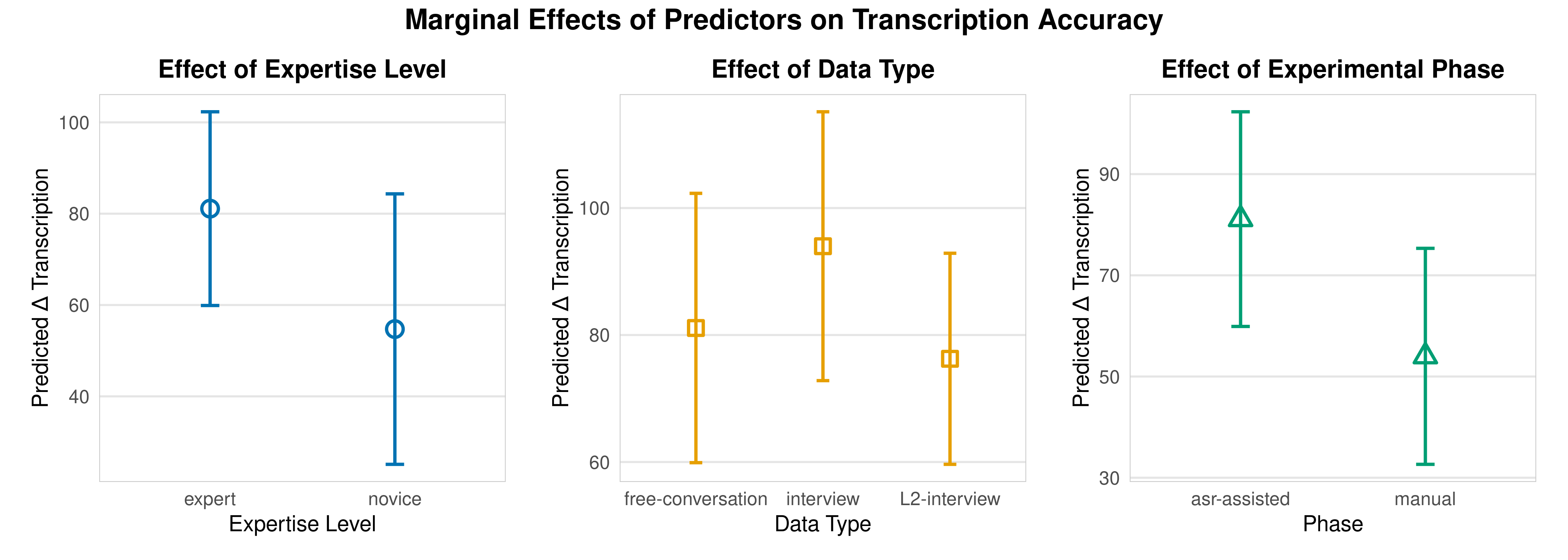}
   \caption{Marginal effects of expertise level (expert/novice), data type (free-conversation/interview/L2-interview), and experimental phase (manual/ASR-assisted) on predicted transcription accuracy}
   \label{fig:effects}
\end{figure*}

\onecolumn
\begin{sidewaystable}[]
\centering
\resizebox{\textwidth}{!}{%
\begin{tabular}{lrrrrrrrr}
\hline
\textbf{variable} & \multicolumn{1}{l}{$R^2$} & \multicolumn{1}{l}{\textbf{deviance}} & \multicolumn{1}{l}{\textbf{transcriber (edf)}} & \multicolumn{1}{l}{\textbf{transcriber (p-value)}} & \multicolumn{1}{l}{\textbf{data (L2-interview)}} & \multicolumn{1}{l}{\textbf{data (interview)}} & \multicolumn{1}{l}{\textbf{expertise}} & \multicolumn{1}{l}{\textbf{phase}} \\ \hline
average tokens per TU (delta) & 0,770 & 82,0 & 8,33 & \textbf{0,00} & 0,62 & \textbf{0,00} & 0,92 & \textbf{0,00} \\
average tokens per TU & 0,767 & 81,5 & 8,49 & \textbf{0,00} & 0,95 & \textbf{0,00} & 0,81 & \textbf{0,00} \\
average TU duration & 0,713 & 77,2 & 8,33 & \textbf{0,00} & 0,23 & \textbf{0,00} & 0,66 & \textbf{0,00} \\
number of linguistic tokens & 0,699 & 72,2 & 0,00 & 0,79 & 0,82 & \textbf{0,00} & 0,22 & 0,28 \\
number of TUs & 0,697 & 75,3 & 6,95 & \textbf{0,00} & 0,22 & \textbf{0,00} & 0,71 & 0,26 \\
number of tokens & 0,693 & 71,7 & 0,00 & 0,67 & 0,74 & \textbf{0,00} & 0,21 & 0,30 \\
number of intonation patterns & 0,693 & 71,7 & 0,00 & 0,67 & 0,74 & \textbf{0,00} & 0,21 & 0,30 \\
duration of TUs & 0,690 & 71,4 & 0,00 & 0,68 & \textbf{0,01} & \textbf{0,00} & 0,58 & 0,50 \\
average TU duration (delta) & 0,599 & 67,9 & 8,01 & \textbf{0,00} & 1,00 & \textbf{0,02} & 0,87 & \textbf{0,03} \\
number of overlaps & 0,521 & 60,6 & 5,54 & \textbf{0,01} & \textbf{0,00} & 0,79 & \textbf{0,03} & \textbf{0,06} \\
number of TUs (delta) & 0,470 & 56,8 & 7,02 & \textbf{0,00} & 0,22 & \textbf{0,00} & 0,58 & 0,24 \\
\rowcolor[HTML]{EFEFEF} 
number of overlaps (delta) & 0,449 & 50,4 & 1,46 & 0,31 & \textbf{0,00} & 0,18 & \textbf{0,04} & 0,22 \\
\rowcolor[HTML]{EFEFEF} 
duration of TUs (delta) & 0,427 & 51,8 & 5,25 & \textbf{0,03} & \textbf{0,00} & \textbf{0,00} & 0,13 & 0,32 \\
\rowcolor[HTML]{EFEFEF} 
number of prolongations (delta) & 0,421 & 52,9 & 7,14 & \textbf{0,00} & \textbf{0,07} & \textbf{0,00} & 0,32 & 0,13 \\
\rowcolor[HTML]{EFEFEF} 
number of prolongations & 0,410 & 50,8 & 5,86 & \textbf{0,01} & 0,24 & 0,76 & 0,24 & \textbf{0,11} \\
\rowcolor[HTML]{EFEFEF} 
number of tokens (delta) & 0,332 & 42,5 & 4,05 & \textbf{0,09} & \textbf{0,00} & \textbf{0,00} & \textbf{0,09} & 0,23 \\
\rowcolor[HTML]{EFEFEF} 
number of intonation patterns (delta) & 0,332 & 42,5 & 4,05 & \textbf{0,09} & \textbf{0,00} & \textbf{0,00} & \textbf{0,09} & 0,23 \\
\rowcolor[HTML]{EFEFEF} 
number of linguistic tokens (delta) & 0,316 & 41,5 & 4,45 & \textbf{0,06} & \textbf{0,00} & \textbf{0,00} & 0,16 & 0,23 \\
\rowcolor[HTML]{EFEFEF} 
number of non-verbal-behaviors & 0,199 & 32,5 & 5,21 & \textbf{0,02} & 0,35 & \textbf{0,06} & 0,52 & 0,37 \\
\rowcolor[HTML]{EFEFEF} 
number of non-verbal-behaviors (delta) & 0,136 & 20,3 & 0,00 & 0,46 & \textbf{0,01} & 0,53 & 0,17 & 0,47 \\
\hline
\end{tabular}%
}
\caption{Generalized Additive Mixed Model (GAMM) results.}
\label{tab:models}
\end{sidewaystable}

\end{document}